%% file: main.tex
\documentclass{article}

\usepackage[final]{neurips_2018}

\usepackage[utf8]{inputenc} 
\usepackage[T1]{fontenc}    
\usepackage{url}            
\usepackage{booktabs}       
\usepackage{amsfonts}       
\usepackage{nicefrac}       
\usepackage{microtype}      

\input{packages}
\input{macros}

\title{Visual Object Networks: Image Generation with Disentangled 3D Representation}

\author{
  Jun-Yan Zhu\\
  MIT CSAIL\\
  \And
  Zhoutong Zhang\\
  MIT CSAIL\\
  \And
  Chengkai Zhang\\
  MIT CSAIL\\
  \And
  Jiajun Wu\\
  MIT CSAIL\\
  \AND
  Antonio Torralba\\
  MIT CSAIL\\
  \And
  Joshua B. Tenenbaum\\
  MIT CSAIL\\
  \And
  William T. Freeman\\
  MIT CSAIL, Google\\
}

\begin{document}

\maketitle

\input{text/abstract.tex}


\input{text/intro}

\input{text/related_work}
\input{text/method}
\input{text/experiments}
\input{text/conclusion}
\clearpage
{\small
\bibliographystyle{plainnat}
\bibliography{reference,main}
}

\end{document}

%% file: packages.tex
\usepackage{color,xcolor}
\usepackage{epsfig}
\usepackage{graphicx}

\usepackage{adjustbox}
\usepackage{array}
\usepackage{booktabs}
\usepackage{colortbl}
\usepackage{float,wrapfig}
\usepackage{hhline}
\usepackage{multirow}
\usepackage{subcaption} 

\usepackage{amsmath,amsfonts,amsthm,amssymb}
\usepackage{bm}
\usepackage{nicefrac}
\usepackage{microtype}
\usepackage{dsfont}
\usepackage{changepage}
\usepackage{extramarks}
\usepackage{fancyhdr}
\usepackage{lastpage}
\usepackage{setspace}
\usepackage{soul}
\usepackage{xspace}

\usepackage[pagebackref=true,breaklinks=true,colorlinks,citecolor=gray]{hyperref}
\usepackage{url}

\usepackage{algorithm, algorithmic}
\usepackage{enumerate}
\usepackage{todonotes} 

\usepackage{titlesec}

%% file: macros.tex
\newcolumntype{L}[1]{>{\raggedright\let\newline\\\arraybackslash\hspace{0pt}}m{#1}}
\newcolumntype{C}[1]{>{\centering\let\newline\\\arraybackslash\hspace{0pt}}m{#1}}
\newcolumntype{R}[1]{>{\raggedleft\let\newline\\\arraybackslash\hspace{0pt}}m{#1}}

\newcommand{\tbl}[1]{Table~\ref{#1}}

\newcommand{\ignore}[1]{}
\newcommand{\norm}[1]{\lVert#1\rVert}

\DeclareMathOperator*{\argmin}{arg\,min}
\DeclareMathOperator*{\argmax}{arg\,max}

\def\suppshort{supplement}
\makeatletter
\DeclareRobustCommand\onedot{\futurelet\@let@token\@onedot}
\def\@onedot{\ifx\@let@token.\else.\null\fi\xspace}

 \def\vs{\emph{vs}\onedot}

\makeatother

\definecolor{MyDarkBlue}{rgb}{0,0.08,1}
\definecolor{MyDarkGreen}{rgb}{0.02,0.6,0.02}
\definecolor{MyDarkRed}{rgb}{0.8,0.02,0.02}
\definecolor{MyDarkOrange}{rgb}{0.40,0.2,0.02}
\definecolor{MyPurple}{RGB}{111,0,255}
\definecolor{MyRed}{rgb}{1.0,0.0,0.0}
\definecolor{MyGold}{rgb}{0.75,0.6,0.12}
\definecolor{MyDarkgray}{rgb}{0.66, 0.66, 0.66}

\newcommand{\ours}{visual object networks\xspace}
\newcommand{\oursshort}{VON\xspace}
\newcommand{\oursshorts}{VON\xspace}

\newcommand{\OURS}{Visual Object Networks\xspace}

\newcommand{\reffig}[1]{Figure~\ref{fig:#1}}
\newcommand{\refsec}[1]{Section~\ref{sec:#1}}

\newcommand{\reftbl}[1]{Table~\ref{tbl:#1}}

\newcommand{\refeq}[1]{Eqn.~\ref{eq:#1}}

\newcommand{\lblfig}[1]{\label{fig:#1}}
\newcommand{\lblsec}[1]{\label{sec:#1}}
\newcommand{\lbleq}[1]{\label{eq:#1}}
\newcommand{\lbltbl}[1]{\label{tbl:#1}}

\newcommand{\ignorethis}[1]{}

\newcommand{\zv}{\rvz_{\text{view}}}
\newcommand{\zs}{\rvz_{\text{shape}}}
\newcommand{\zt}{\rvz_{\text{texture}}}

\newcommand{\Gs}{G_{\text{shape}}}
\newcommand{\Ds}{D_{\text{shape}}}
\newcommand{\Gt}{G_{\text{texture}}}
\newcommand{\Dt}{D_{\text{texture}}}
\newcommand{\Di}{D_{\text{image}}}
\newcommand{\Dtf}{D_{\text{2.5D}}}
\newcommand{\Et}{E_{\text{texture}}}
\newcommand{\Etf}{E_{\text{2.5D}}}
\newcommand{\depth}{\text{depth}}
\newcommand{\mask}{\text{mask}}

\def\vone{{\bm{1}}}
\newcommand{\voxel}{\rvv}
\newcommand{\voxelt}{\tilde{\rvv}}
\newcommand{\image}{\rvx}
\newcommand{\tf}{\rvv_{\text{2.5D}}}
\newcommand{\DP}{\mathcal{P}}

\newcommand{\Ls}{\mathcal{L}_{\text{shape}}}
\newcommand{\Lt}{\mathcal{L}_{\text{texture}}}
\newcommand{\Li}{\mathcal{L}_{\text{image}}}
\newcommand{\Ltf}{\mathcal{L}_{\text{2.5D}}}
\newcommand{\GAN}{^{\text{GAN}}}
\newcommand{\lams}{\lambda_{\text{shape}}}

\newcommand{\fid}{Fréchet Inception Distance\xspace}
\newcommand{\fids}{Fréchet Inception Distances\xspace}

\newcommand{\myparagraph}[1]{\vspace{-0pt}\paragraph{#1}}

\def\rvv{{\mathbf{v}}}

\def\rvx{{\mathbf{x}}}

\def\rvz{{\mathbf{z}}}

%% file: text/abstract.tex
\begin{abstract}
Recent progress in deep generative models has led to tremendous breakthroughs in image generation. However, while existing models can synthesize photorealistic images, they lack an understanding of our underlying 3D world. We present a new generative model, \emph{\OURS (\oursshorts)}, synthesizing natural images of objects with a disentangled 3D representation. Inspired by classic graphics rendering pipelines, we unravel our image formation process into three conditionally independent factors---shape, viewpoint, and texture---and present an end-to-end adversarial learning framework that jointly models 3D shapes and 2D images. Our model first learns to synthesize 3D shapes that are indistinguishable from real shapes. It then renders the object's 2.5D sketches (i.e., silhouette and depth map) from its shape under a sampled viewpoint. Finally, it learns to add realistic texture to these 2.5D sketches to generate natural images. The \oursshort not only generates images that are more realistic than state-of-the-art 2D image synthesis methods, but also enables many 3D operations such as changing the viewpoint of a generated image,  editing of shape and texture, linear interpolation in texture and shape space, and transferring appearance across different objects and viewpoints. 

\end{abstract}

%% file: text/intro.tex
\section{Introduction}
\lblsec{intro}

Modern deep generative models learn to synthesize realistic images. \reffig{teaser}a shows several cars generated by a recent model~\citep{Gulrajani2017Improved}. However, most methods have only focused on generating images in 2D, ignoring the 3D nature of the world. As a result, they are unable to answer some questions that would be effortless for a human, for example: what will a car look like from a different angle? What if we apply its texture to a truck? Can we mix different 3D designs? 
Therefore, a 2D-only perspective inevitably limits a model’s practical application in fields such as robotics, virtual reality, and gaming.

In this paper, we present an end-to-end generative model that jointly synthesizes 3D shapes and 2D images via a disentangled object representation. Specifically, we decompose our image generation model into three conditionally independent factors: shape, viewpoint, and texture, borrowing ideas from classic graphics rendering engines~\citep{kajiya1986rendering}. Our model first learns to synthesize 3D shapes that are indistinguishable from real shapes. It then computes its 2.5D sketches~\citep{Barrow1978Recovering,Marr1982Vision} with a differentiable projection module from a sampled viewpoint. Finally, it learns to add diverse, realistic texture to 2.5D sketches and produce 2D images that are indistinguishable from real photos. We call our model \OURS (\oursshorts).

Wiring in conditional independence reduces our need for densely annotated data: unlike classic morphable face models~\citep{blanz1999morphable}, our training does not require paired data between 2D images and 3D shapes, nor dense correspondence annotations in 3D data. This advantage allows us to leverage both 2D image datasets and 3D shape collections~\citep{Chang2015Shapenet} and to synthesize objects of diverse shapes and texture.

Through extensive experiments, we show that \oursshorts produce more realistic image samples than recent 2D deep generative models. We also demonstrate many 3D applications that are enabled by our disentangled representation, including rotating an object, adjusting object shape and texture, interpolating between two objects in texture and shape space independently, and transferring the appearance of a real image to new objects and viewpoints.

\input{figText/teaser.tex}

%% file: figText/teaser.tex
\begin{figure}[t]
  \centering
  \includegraphics[width=\linewidth]{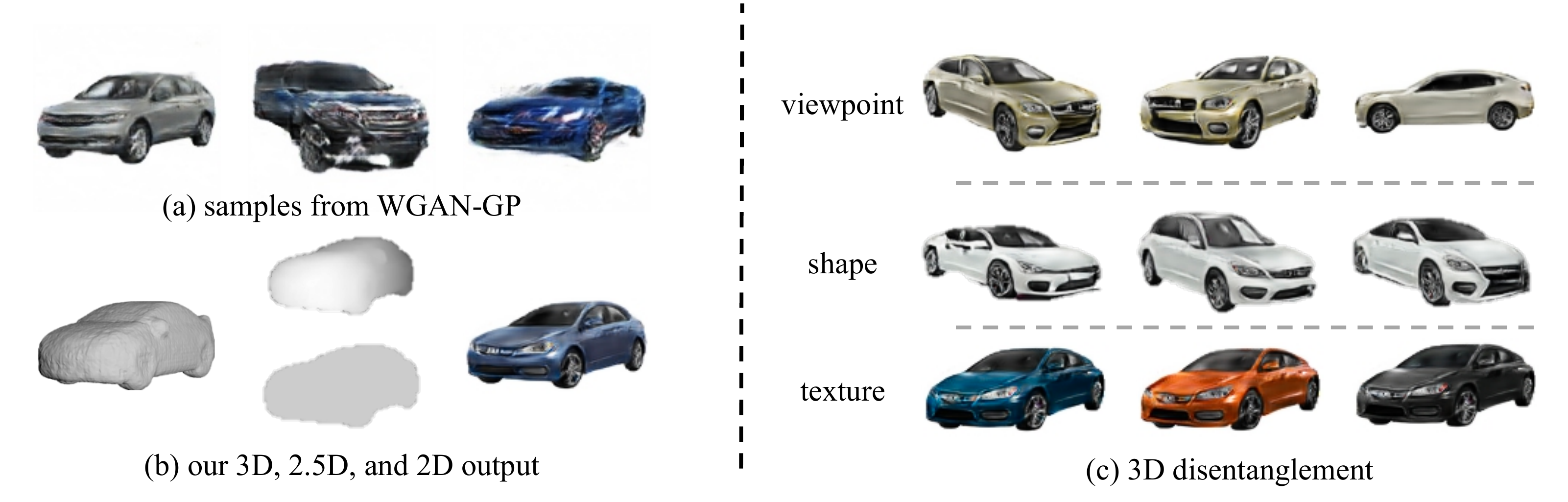}
  \caption{Previous 2D GANs vs. \OURS (\oursshorts). (a) Typical examples produced by a recent GAN model~\citep{Gulrajani2017Improved}. (b) Our model produces three outputs:  a 3D shape,  its 2.5D projection given a viewpoint, and a final image with realistic texture. (c) Given this disentangled 3D representation, our method allows several 3D applications including changing viewpoint and editing shape or texture independently. Please see our \href{https://github.com/junyanz/VON}{code} and \href{http://von.csail.mit.edu}{website} for more details.}
  \lblfig{teaser}
   \vspace{-13pt}
\end{figure}

%% file: text/related_work.tex
\section{Related Work}
\lblsec{related}
\paragraph{GANs for 2D image synthesis.}
Since the invention of Generative Adversarial Nets (GANs)~\citep{Goodfellow2014Generative}, many researchers have adopted adversarial learning  for various image synthesis tasks, ranging from image generation~\citep{Radford2016Unsupervised,arjovsky2017wasserstein,karras2018progressive}, image-to-image translation \citep{Isola2017Image,Zhu2017Unpaired}, text-to-image synthesis~\citep{zhang2017stackgan,reed2016generative}, and interactive image editing~\citep{Zhu2016Generative,wang2018high}, to classic vision and graphics tasks such as inpainting~\citep{pathak2016context} and super-resolution~\citep{Ledig2017Photo}.
Despite the tremendous progress made on 2D image synthesis, most of the above methods operate on 2D space, ignoring the 3D nature of our physical world. As a result, the lack of 3D structure inevitably limits some practical applications of these generative models. In contrast, we present an image synthesis method powered by a disentangled 3D representation.It allows a user to change the viewpoint easily, as well as to edit the object's shape or texture independently. \cite{Dosovitskiy2015Learning} used supervised CNNs for generating synthetic images given object style, viewpoint, and color. We differ in that our aim is to produce objects with 3D geometry and natural texture without using labelled data.

\myparagraph{3D shape generation.}
There has been an increasing interest in synthesizing 3D shapes with deep generative models, especially GANs. Popular representations include voxels~\citep{Wu2016Learning}, point clouds~\citep{gadelha2017shape,Achlioptas2018Learning}, and octave trees~\citep{Tatarchenko2017Octree}. Other methods learn 3D shape priors from 2D images~\citep{Rezende2016Unsupervised,gadelha20173d}. Recent work also explored 3D shape completion from partial scans with deep generative models~\citep{Dai2017Shape,Wang2018Shape,wu2018learning}, including generalization to unseen object categories~\citep{Zhang2018Learning}. Unlike prior methods that only synthesize untextured 3D shapes, our method learns to produce both realistic shapes and images. Recent and concurrent work has learned to infer both texture and 3D shapes from 2D images, represented as parametrized meshes~\citep{Kanazawa2018Learning}, point clouds~\citep{Tatarchenko2016Multi}, or colored voxels~\citep{Tulsiani2017Learning,Sun2018Im2Avatar}. While they focus on 3D reconstruction, we aim to learn an unconditional generative model of shapes and images with disentangled representations of object texture, shape and pose. 

\myparagraph{Inverse graphics.}
Motivated by the philosophy of ``vision as inverse graphics''~\citep{Yuille2006Vision,Bever2010Analysis}, researchers have made much progress in recent years on learning to invert graphics engines, many with deep neural networks~\citep{Kulkarni2015Deep,Yang2015Weakly,Kulkarni2015Picture,tung2017adversarial,shu2017neural}. In particular, \cite{Kulkarni2015Deep} proposed a convolutional inverse graphics network. Given an image of a face, the network learns to infer properties such as pose and lighting.  \cite{tung2017adversarial} extended inverse graphics networks with adversarial learning. \cite{Wu2017MarrNet,wu2018learning} inferred 3D shapes from a 2D  image via 2.5D sketches and learned shape priors. Here we focus on a complementary problem---learning generative graphics networks via the idea of ``graphics as inverse vision''. In particular, we learn our generative model with  recognition models that recover 2.5D sketches from generated images.

%% file: text/method.tex
\section{Formulation}
\lblsec{method}

\input{figText/overview.tex}

Our goal is to learn an (implicit) generative model that can sample an image $x \in \mathds{R}^{H\!\times W\!\times 3}$ from three factors: a shape code $\zs$, a viewpoint code $\zv$, and a texture code $\zt$. The texture code describes the appearance of the object,  which accounts for the object's albedo, reflectance, and environment illumination. These three factors are disentangled,  conditionally independent from each other. Our model is category-specific, as the visual appearance of an object depends on the class. We further assume that all the codes lie in their own low-dimensional spaces. During training, we are given a 3D shape collection $\{\voxel_i\}_i^N$, where $\voxel_i \in \mathds{R}^{W\!\times W\!\times W}$ is a binary voxel grid, and a 2D image collection $\{\image_j\}_j^M$, where $\image_j \in \mathds{R}^{H\!\times W\!\times 3}$.  Our model training requires \textit{no} alignment between 3D and 2D data. We assume that every training image has a clean background and only contains the object of interest. This assumption makes our model focus on generating realistic images of the objects instead of complex backgrounds.

\reffig{overview} illustrates our model. First, we learn a 3D shape generation network that produces realistic voxels $\voxel=\Gs(\zs)$ given a shape code $\zs$ (\refsec{learn_shape}). We then develop a differentiable projection module $\DP$ that projects a 3D voxel grid $\voxel$ into 2.5D sketches via $\tf = \DP(\voxel, \zv)$, given a particular viewpoint  $\zv$ (\refsec{project_2.5d}). Next, we learn to produce a final image given the 2.5D sketches $\tf$ and a randomly sampled texture code $\zt$, using our texture synthesis network $\image =\Gt(\tf, \zt)$ in \refsec{learn_texture}.
\refsec{full} summarizes our full model and \refsec{implement} includes implementation details. Our entire model is differentiable and can be trained end-to-end.

During testing, we sample an image $x=\Gt(\DP(\Gs(\zs), \zv), \zt)$ from latent codes $(\zs, \zv, \zt)$ via our shape network $\Gs$,  texture network $\Gt$,  and projection module $\DP$.

\subsection{Learning 3D Shape Priors}
\lblsec{learn_shape}
Our first step is to learn a category-specific 3D shape prior from large shape collections~\citep{Chang2015Shapenet}. This prior depends on the object class but is conditionally independent of other factors such as viewpoint and texture.  To model the 3D shape prior and generate realistic shapes, we adopt the 3D Generative Adversarial Networks recently proposed by ~\citet{Wu2016Learning}. 

Consider a voxelized 3D object collection $\{\voxel_i\}_i^N$, where $\voxel_i \in \mathds{R}^{W\!\times W\!\times W}$. We learn a generator $\Gs$ to map a shape code $\zs$, randomly sampled from a Gaussian distribution, to a ${W\!\times W\!\times W}$ voxel grid. Simultaneously, we train a 3D discriminator $\Ds$ to classify a shape as real or generated. Both discriminator and generator contain fully volumetric convolutional and deconvolutional layers. We find that the original 3D-GAN~\citep{Wu2016Learning} sometimes suffers from mode collapse. To improve the quality and diversity of the results, we use the Wasserstein distance of WGAN-GP~\citep{arjovsky2017wasserstein,Gulrajani2017Improved}. Formally, we play the following minimax two-player game between $\Gs$ and $\Ds$: $\min_{\Gs}\max_{\Ds} \Ls^{\text{GAN}}$\footnote{For notation simplicity, we denote  $ \mathbb{E}_{\voxel} \triangleq \mathbb{E}_{\voxel \sim p_{\text{data}}(\voxel)}$ and $\mathbb{E}_{\zs} \triangleq \mathbb{E}_{\zs \sim p_{\text{data}}(\zs)}$.},
where
\begin{equation}
 \Ls= \mathbb{E}_{\voxel} [\Ds(\voxel)] -\mathbb{E}_{\zs}[\Ds(\Gs(\zs)].
    \lbleq{3DGAN_G}
\end{equation}
To enforce the Lipschitz constraint in Wasserstein GANs~\citep{arjovsky2017wasserstein}, we add a gradient-penalty loss $\lambda_{\text{GP}} \mathbb{E}_{\voxelt}[(\nabla_{\voxelt}\Ds(\voxelt)-1)^2]$ to \refeq{3DGAN_G}, where $\voxelt$ is a randomly sampled point along the straight line between a real shape and a generated shape, and $\lambda_{\text{GP}}$ controls the capacity of $\Ds$. Since binary data is often challenging to model using GANs, we also experiment with distance function (DF) representation~\citep{curless1996volumetric}, which is continuous on the 3D voxel space.  See \refsec{expr_voxel} for quantitative evaluations. 

\subsection{Generating 2.5D Sketches}
\lblsec{project_2.5d}
Given a synthesized voxelized shape $v=\Gs(\zs)$, how can we connect it to a 2D image? Inspired by recent work on 3D reconstruction~\citep{Wu2017MarrNet}, we use 2.5D sketches~\citep{Barrow1978Recovering,Marr1982Vision} to bridge the gap between 3D and 2D. This intermediate representation provides three main advantages. First, generating 2.5D sketches from a 3D voxel grid is straightforward, as the projection is differentiable with respect to both the input shape and the viewpoint. Second, 2D image synthesis from a 2.5D sketch can be cast as an image-to-image translation problem~\citep{Isola2017Image}, where existing methods have achieved successes even without paired data~\citep{Zhu2017Unpaired}. Third, compared with alternative approaches such as colored voxels, our method enables generating images at a higher resolution.

Here we describe our differentiable module for projecting voxels into 2.5D sketches. The inputs to this module are the camera parameters and 3D voxels. The value of each voxel stores the probability of it being present. To render the 2.5D sketches from the voxels under a perspective camera, we first generate a collection of rays, each originating from the camera's center and going through a pixel's center in the image plane. To render the 2.5D sketches, we need to calculate whether a given ray would hit the voxels, and if so, the corresponding depth value of that ray. To this end, we first sample a collection of points at evenly spaced depth along each ray. Next, for each point, we calculate the probability of hitting the input voxels using a differentiable trilinear interpolation~\citep{jaderberg2015spatial} of the input voxels. Similar to \cite{Tulsiani2017Learning}, we then calculate the expectation of visibility and depth along each ray. Specifically, given a ray $R$ with $N$ samples $R_1$, $R_2$, ... , $R_N$ along its path, we calculate the visibility (silhouette)  as the expectation of the ray hitting the voxels: $\sum_{j=1}^N \prod_{k=1}^{j-1} (1-R_k)R_j$. Similarly,
 the expected depth can be calculated as $\sum_{j=1}^N d_j \prod_{k=1}^{j-1} (1-R_k) R_j$, where $d_j$ is the depth of the sample $R_j$. This process is fully differentiable since the gradients can be back-propagated through both the expectation calculation and the trilinear interpolation.

\myparagraph{Viewpoint estimation.} Our two-dimensional viewpoint code $\zv$ encodes camera elevation and azimuth. We sample $\zv$
 from an empirical distribution $p_{\text{data}}(\zv)$ of the camera poses from the training images.  To estimate $p_{\text{data}}(\zv)$, we first render the silhouettes of several candidate 3D models under uniformly sampled camera poses. For each input image, we compare its silhouette to the rendered 2D views and choose the pose with the largest Intersection-over-Union value. More details can be found in the \suppshort.

\subsection{Learning 2D Texture Priors}
\lblsec{learn_texture}
Next, we learn to synthesize realistic 2D images given projected 2.5D sketches that encode both the viewpoint and the object shape. In particular, we learn a texture network $\Gt$ that takes a randomly sampled texture code $\zt$ and the projected 2.5D sketches $\tf$ as input, and produces a 2D image $\image=\Gt(\tf, \zt)$. This texture network needs to model both object texture and environment illumination, as well as the differentiable rendering equation~\citep{kajiya1986rendering}. Fortunately, this mapping problem can be cast as an unpaired image-to-image translation problem~\citep{Zhu2017Unpaired,yi2017dualgan,liu2017unsupervised}. We adopt recently proposed cycle-consistent adversarial networks (CycleGAN)~\citep{Zhu2017Unpaired} as our baseline. Later, we relax the one-to-one mapping restriction in CycleGAN to handle one-to-many mappings from 2.5D sketches to 2D images.

Here we introduce two encoders $\Et$ and $\Etf$ to estimate a texture code $\zt$ and 2.5D sketches $\tf$ from a real image $\image$. We train $\Gt$, $\Et$, and $\Etf$ jointly with adversarial losses~\citep{Goodfellow2014Generative} and cycle-consistency losses~\citep{Zhu2017Unpaired,yi2017dualgan}. We use the following adversarial loss on the final generated image:
\begin{equation}   
\Li\GAN = \mathbb{E}_{\image} [\log \Di(\image)] + \mathbb{E}_{(\tf, \zt)}[\log (1-\Di(\Gt(\tf, \zt))], 
    \lbleq{GAN_G}
\end{equation}
where $\Di$ learns to classify real and generated images. We apply the same adversarial loss for 2.5D sketches $\tf$:
\begin{equation}
 \Ltf\GAN = \mathbb{E}_{\tf} [\log \Dtf(\tf)] + \mathbb{E}_{\image}[\log (1-\Dtf(\Etf(\image))], 
    \lbleq{GAN_E}
\end{equation}
where $\Dtf$ aims to distinguish between 2.5D sketches $\tf$ and estimated 2.5D sketches $\Etf(\image)$ from a real 2D image. We further use cycle-consistency losses~\citep{Zhu2017Unpaired} to enforce the bijective relationship between the two domains: 
\begin{align}
    &\mathcal{L}^{\text{cyc}}_{\text{2.5D}} = \lambda^{\text{cyc}}_{\text{2.5D}} \mathbb{E}_{(\tf, \zt)}\left[\norm{\Etf(\Gt(\tf, \zt))-\tf}_1\right] \nonumber \\
    \text{and} \quad &\mathcal{L}^{\text{cyc}}_{\text{image}} = \lambda^{\text{cyc}}_{\text{image}} \mathbb{E}_{\image}\left[\norm{\Gt(\Etf(\image), \Et(\image))-\image}_1\right],
    \lbleq{image_cycle}
\end{align}
where $\lambda^{\text{cyc}}_{\text{image}}$ and $\lambda^{\text{cyc}}_{\text{2.5D}}$ control the importance of each cycle loss. 
TThe texture encoder $\Et$ and 2.5D sketch encoder $\Etf$ serve as recognition models that recover the texture and 2.5D representation from a 2D image. 

\myparagraph{One-to-many mappings.} Prior studies~\citep{Isola2016Learning,Mathieu2016Deep} have found that latent codes are often ignored in conditional image generation  due to the assumption of a one-to-one mapping; vanilla CycleGAN also suffers from this problem based on our experiments. To address this, we introduce a latent space cycle-consistency loss to encourage $\Gt$ to use the texture code $\zt$: 
\begin{equation}
\Lt^{\text{cyc}} = \lambda^{\text{cyc}}_{\text{texture}} \mathbb{E}_{(\tf, \zt)}[\norm{\Et(\Gt(\tf, \zt))-\zt}_1],
    \lbleq{texture_cycle}
\end{equation}
where $\lambda^{\text{cyc}}_{\text{texture}}$ controls its importance. 
Finally, to allow sampling at test time, we add a Kullback–Leibler (KL) loss on the $z$ space to force $\Et(\image)$ to be close to a Gaussian distribution:
\begin{equation}
\mathcal{L}_{\text{KL}} =\lambda_{\text{KL}}  \mathbb{E}_{\image} \left[\mathcal{D}_{\text{KL}}(\Et(\image) ||\mathcal{N}(0,I))\right],    
\end{equation}
where $\mathcal{D}_{\text{KL}}(p||q)=-\int_{\rvz}  p(\rvz) \log\frac{p(\rvz)}{q(\rvz)}dz$ and $\lambda_{\text{KL}}$ is its weight.  We write the final texture loss as
\begin{equation}
\Lt = \underbrace{\Li\GAN + \Ltf\GAN}_{\text{Adversarial losses}} + \underbrace{\mathcal{L}^{\text{cyc}}_{\text{image}} +\mathcal{L}^{\text{cyc}}_{\text{2.5D}}+\Lt^{\text{cyc}}}_{\text{Cycle-consistency losses}} +  \underbrace{\mathcal{L}_{\text{KL}}}_{\text{KL loss}}.
\lbleq{texture}
\end{equation}
Note that the latent space reconstruction loss $\Lt^{\text{cyc}}$ has been explored in unconditional GANs~\citep{chen2016InfoGAN} and image-to-image translation~\citep{Zhu2017Toward,almahairi2018augmented}. Here we use this loss to learn one-to-many mappings from unpaired data.

\subsection{Our Full Model}
\lblsec{full}

Our full objective is
\begin{equation}
\argmin_{(\Gs, \Gt, \Etf, \Et)} \argmax_{(\Ds, \Dt, \Dtf)} \lams \Ls + \Lt, 
\lbleq{full}
\end{equation}
where $\lams$ controls the relative weight of shape and texture loss functions. We compare our \ours against 2D deep generative models in \refsec{eva}.

\subsection{Implementation Details}
\lblsec{implement}
\myparagraph{Shape networks.} For shape generation, we adopt the 3D-GAN architecture from~\cite{Wu2016Learning}. In particular, the discriminator $\Ds$ contains $6$ volumetric convolutional layers and the generator $\Gs$ contains $6$ strided-convolutional layers. We remove the batch normalization layers~\citep{ioffe2015batch} in the $\Gs$ as suggested by the WGAN-GP paper~\citep{Gulrajani2017Improved}.

\myparagraph{Texture networks.} For texture generation, we use the ResNet encoder-decoder~\citep{Zhu2017Unpaired,Huang2018Multimodal} and concatenate the texture code $\zt$ to intermediate layers in the encoder. For the discriminator, we use two-scale PatchGAN classifiers~\citep{Isola2017Image,Zhu2017Unpaired} to classify overlapping patches as real or fake.  We use a least square objective as in LS-GAN~\citep{mao2017least} for stable training. We use ResNet encoders~\citep{He2015Deep} for our $\Et$ and $\Etf$.

\myparagraph{Differentiable projection module.} We assume the camera is at a fixed distance of $2$m to the object's center and use a focal length of $50$mm ($35$mm film equivalent). The resolution of the rendered sketches are $128\times128$, and we sample $128$ points evenly along each camera ray. We also assume no in-plane rotation, that is, no tilting in the image plane. We implement a custom CUDA kernel for sampling along the projection rays and calculating the stop probabilities.

\myparagraph{Training details.} We train our models on $128\times128\times128$  shapes (voxels or distance function) and $128\times128\times3$  images. During training, we first train the shape generator $\Gs$ on 3D shape collections and then train the texture generator $\Gt$ given ground truth 3D shape data and image data. Finally, we fine-tune both modules together. We sample the shape code $\zs$ and texture code $\zt$ from the standard Gaussian distribution $N(0, I)$, with the code length $|\zs|=200$ and $|\zt|=8$. The entire training usually takes two to three days. For hyperparameters, we set  $\lambda_{\text{KL}}=0.05$, $\lambda_{\text{GP}}=10$, $\lambda^{\text{cyc}}_{\text{image}}=10$, $\lambda^{\text{cyc}}_{\text{2.5D}}=25$, $\lambda^{\text{cyc}}_{\text{texture}}=1$, and $\lambda_{\text{shape}}=0.05$. We use the Adam solver~\citep{Kingma2015Adam} with a learning rate of $0.0002$ for shape generation and  $0.0001$ for texture generation.

We observe that the texture generator $\Gt$ sometimes introduces the undesirable effect of changing the shape of the silhouette when rendering 2.5D sketches $\tf$ (i.e., $\depth$ and $\mask$). To address this issue,  we explicitly mask the generated 2D images with the silhouette from $\tf$: i.e.,  $\Gt(\tf, \zt) = \mask \cdot g_{\text{texture}}(\depth)  + (1-\mask)\cdot \vone$, where $\vone$ is the background white color and the generator $g_{\text{texture}}$ synthesizes an image given a depth map. Similarly, we reformulate $\Etf(x) = (e_{\text{2.5D}}(\image)\cdot \mask_{\text{gt}}, \mask_{\text{gt}})$, where the encoder $e_{\text{2.5D}}$  only predicts $\depth$, and the input object mask is used. In addition, we add a small mask consistency loss $||e_{\text{2.5D}}(\image)-\mask_{\text{gt}}||_1$to encourage the predicted depth map to be consistent with the the object mask. As our training images have clean background, we can estimate the object mask with a simple threshold. 

%% file: figText/overview.tex
\begin{figure}[t]
  \centering
  \includegraphics[width=\linewidth]{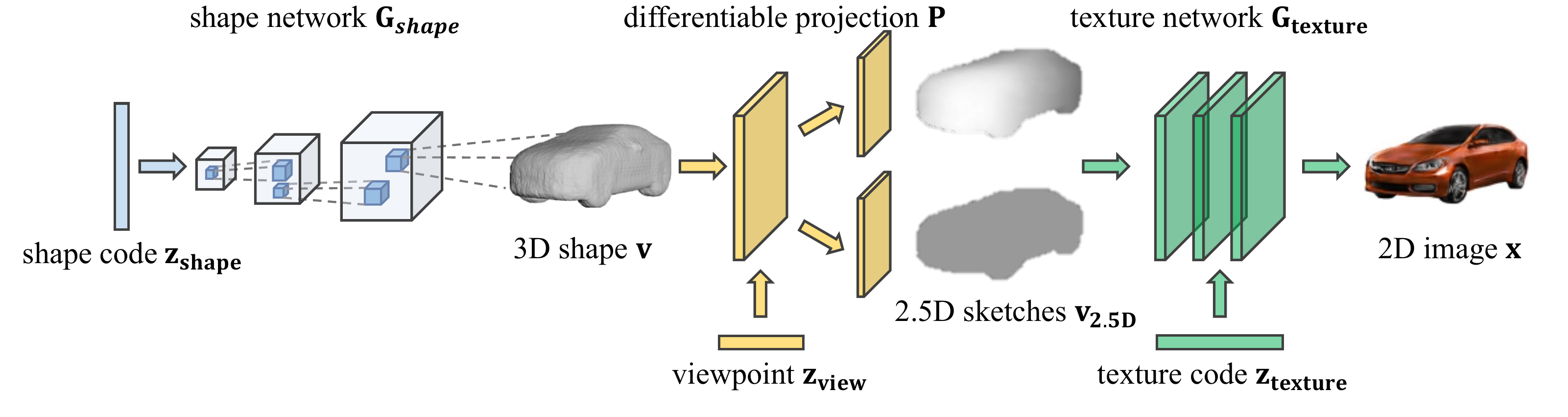}
  \caption{{\bf Our image formation model.} We first learn a shape generative adversarial network $\Gs$ that maps a randomly sampled shape code $\zs$ to a voxel grid $\voxel$. Given a sampled viewpoint $\zv$, we project $\voxel$ to 2.5D sketches $\tf$ with our differentiable projection module $\DP$. The 2.5D sketches $\tf$ include both the object's depth and silhouette, which help to bridge 3D and 2D data. Finally, we learn a texture network $\image=\Gt(\tf, \zt)$ to add realistic, diverse texture to these 2.5D sketches, so that generated 2D images cannot be distinguished from real images by an image discriminator. The model is fully differentiable and trained end-to-end on both 2D and 3D data. }
  \lblfig{overview}
  \vspace{-5pt}
\end{figure}

%% file: text/experiments.tex
\vspace{-10pt}
\section{Experiments}
\vspace{-5pt}
\input{figText/samples.tex}
\lblsec{expr}
We first compare our \ours (\oursshorts) against recent 2D GAN variants on two datasets. We evaluate the results using both a quantitative metric and a qualitative human perception study. We then perform an ablation study on the objective functions of our shape generation network. Finally, we demonstrate several applications enabled by our disentangled 3D representation. The full results and datasets can be found at our \href{http://von.csail.mit.edu}{website}. Please find our implementation at \href{https://github.com/junyanz/VON}{GitHub}. 
\vspace{-8pt}
\subsection{Evaluations}
\lblsec{eva}
\vspace{-3pt}

\myparagraph{Datasets.} We use ShapeNet~\citep{Chang2015Shapenet} for learning to generate 3D shapes. ShapeNet is a large shape repository of $55$ object categories. Here we use the chair and car categories, which has $6,777$ and $3,513$ CAD models respectively. For 2D datasets, we use the recently released Pix3D dataset to obtain $1,515$ RGB images of chairs alongside with their silhouettes~\citep{sun2018pix3d}, with an addition of $448$ clean background images crawled from Google image search. We also crawled $2,605$ images of cars. 
\vspace{-8pt}
\myparagraph{Baselines}
We compare our method to three popular GAN variants commonly used in the literature: DCGAN with the standard cross-entropy loss~\citep{Goodfellow2014Generative,Radford2016Unsupervised}, LSGAN~\citep{mao2017least}, and WGAN-GP~\citep{Gulrajani2017Improved}. We use the same DCGAN-like generator and discriminator architectures for all three GAN models. For WGAN-GP, we replace the BatchNorm by InstanceNorm~\citep{Ulyanov2016Instance} in the discriminator, and we train the discriminator $5$ times per generator iteration.

\input{figText/tbl_fid2d.tex}

\input{figText/tbl_fid3d.tex}

\myparagraph{Metrics.} To evaluate the image generation models, we calculate the \fid between generated images and real images, a metric highly correlated to human perception~\citep{heusel2017gans,lucic2017gans}. Each set of images are fed to the Inception network~\citep{Szegedy2015Going} trained on ImageNet~\citep{Deng2009Imagenet}, and the features from the layer before the last fully-connected layer are used to calculate the \fid. 

Second, we sample $200$ pairs of generated images from the \oursshort and the state-of-the-art models (DCGAN, LSGAN, and WGAN-GP), and show each pair to five subjects on Amazon MTurk. The subjects are asked to choose a more realistic result within the pair.  

\myparagraph{Results}

Our \oursshort consistently outperforms the 2D generative models. In particular, \tbl{tbl:fid2d} shows that our results have the smallest \fid; in \tbl{tbl:user2d},  $74\%-85\%$ of the responses preferred our results. This performance gain demonstrates that the learned 3D prior helps synthesize more realistic images. See \reffig{samples} for a qualitative comparison between these methods.

\myparagraph{Analysis of shape generation.} For shape generation, we compare our method against the prior 3D-GAN work by ~\cite{Wu2016Learning} on both voxel grids and distance function representation. 3D-GAN uses the same architecture but trained with a cross-entropy loss. We evaluate the shape generation models using the \fid (FID) between the generated and real shapes. To extract statistics for each set of generated/real shapes, we train ResNet-based 3D shape classifiers~\citep{He2015Deep} on all $55$ classes of shapes from ShapeNet; classifiers are trained separately on voxels and distance function representations. We extract the features from the layer before the last fully-connected layer. \reftbl{fid3d} shows that our method achieves better results regarding FID. \reffig{shape_comp}a shows that the Wasserstein distance increases the quality of the results. As we use different classifiers for voxels and distance functions, the \fid is not comparable across representations.

\lblsec{expr_voxel}
\subsection{Applications}
\lblsec{app}
\input{figText/applications.tex}
\vspace{-5pt}

We apply our \ours to several 3D manipulation applications, not possible by previous 2D generative models~\citep{Goodfellow2014Generative,Kingma2014Auto}.

\myparagraph{Changing viewpoints.}  As our \oursshort first produces a 3D shape, we can project the shape to the image plane given different viewpoints $\zv$ while keeping the same shape and texture code. \reffig{teaser}c and~\reffig{app}a show a few examples. 

\myparagraph{Shape and texture editing.}  With our learned disentangled 3D representation, we can easily change only the shape code or the texture code, which allows us to edit the shape and texture separately. See \reffig{teaser}c and~\reffig{app}a for a few examples.

\myparagraph{Disentangled interpolation.} Given our disentangled 3D representation, we can choose to interpolate between two objects in different ways. For example, we can interpolate objects in shape space $\alpha \zs^1 + (1-\alpha) \zs^2$ with the same texture, or in the texture space $\alpha \zt^1 + (1-\alpha) \zt^2$ with the same shape, or both, where $\alpha \in [0, 1]$. \reffig{app}c shows linear interpolations in the latent space. 

\input{figText/transfer.tex}

\myparagraph{Example-based texture transfer.}  We can infer the texture code $\zt$ from a real image $\image$ with the texture encoder $\zt=\Et(\image)$, and apply the code to new shapes. \reffig{transfer} shows texture transfer results on cars and chairs using real images and generated shapes.

%% file: figText/samples.tex
\begin{figure}[t]
  \centering
   \includegraphics[width=\linewidth]{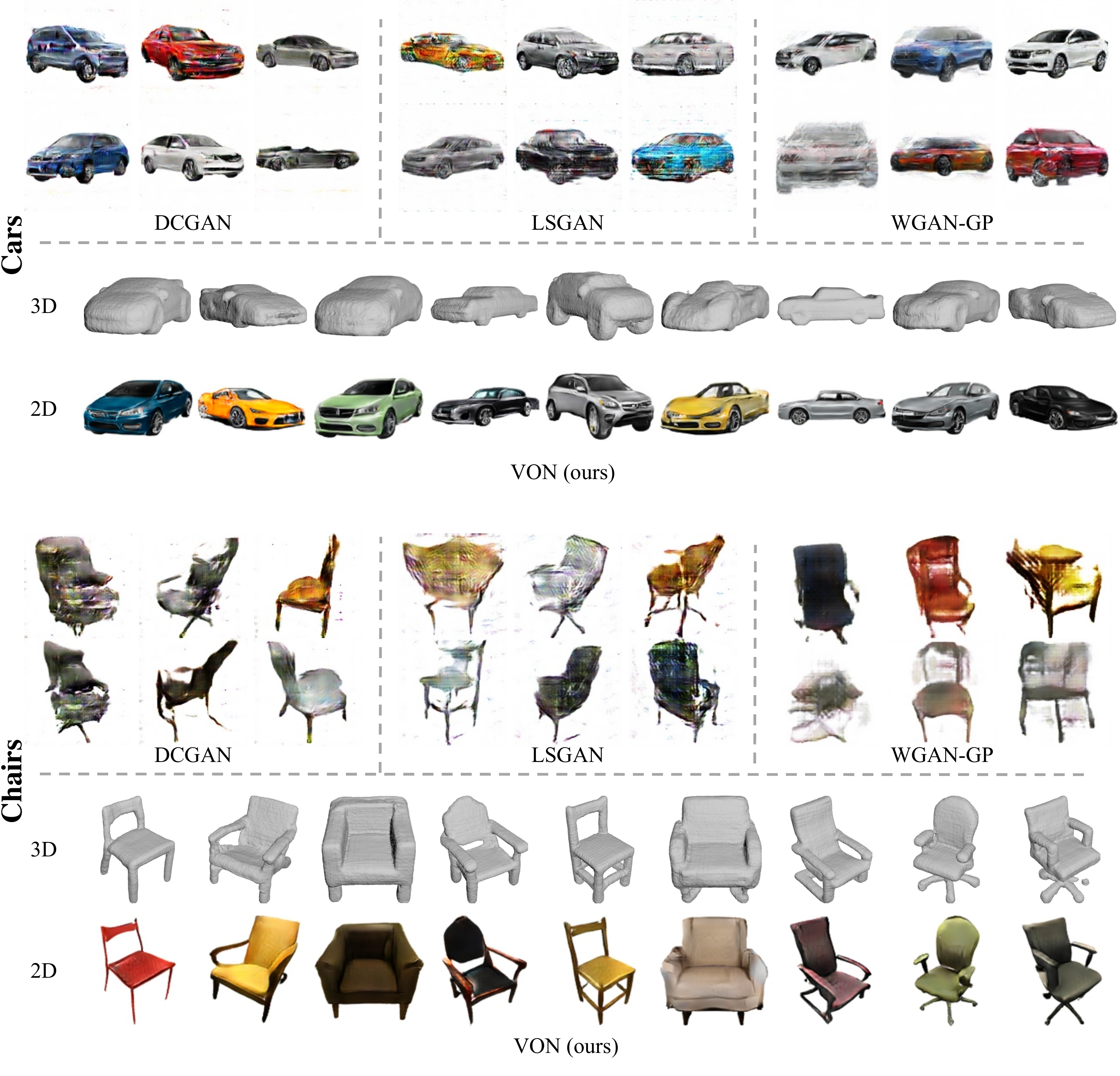}
   \vspace{-5pt}
  \caption{Qualitative comparisons between 2D GAN models and \oursshort: we show samples from DCGAN~\citep{Radford2016Unsupervised}, LSGAN~\citep{mao2017least}, WGAN-GP~\citep{Gulrajani2017Improved}, and our \oursshort. For our method, we show both 3D shapes and 2D images. Note that \oursshort is trained on unpaired 3D shapes and 2D images, while DCGAN, LSGAN and WGAN-GP are trained only on 2D images. The learned 3D prior helps our model produce better samples. (Top: cars; bottom: chairs.)}
  \lblfig{samples}
  \vspace{-15pt}
\end{figure}

%% file: figText/tbl_fid2d.tex
\begin{table}[t]
\parbox{.47\linewidth}{
 	\centering
    \begin{tabular}{lcc}
    \toprule
     & Car & Chair \\
    \midrule
    DCGAN & 130.5 & 225.0 \\
    LSGAN & 171.4 & 225.3 \\
    WGAN-GP &  123.4 & 184.9 \\
    \oursshort (voxels)  & {\bf 81.6 }  &  58.0 \\
    \oursshort (DF)  & {\bf 83.3 } &  {\bf  51.8}
\\
    \bottomrule
    \end{tabular}
    \vspace{5pt}
    \caption{\fids~\citep{heusel2017gans} between real images and images generated by DCGAN, LSGAN, WGAN-GP, our \oursshort (voxels), and our \oursshort (DF). DF denotes distance function representations. }
    \label{tbl:fid2d}
    \vspace{-20pt}
    }
\hfill
\parbox{.47\linewidth}{
 	\centering
    \begin{tabular}{lcc}
    \toprule
     & Car & Chair \\
    \midrule
    DCGAN &  72.2\% & 90.3\% \\
    LSGAN &  78.7\%  &  92.4\%\\
    WGAN-GP   & 63.0 \%   & 89.1\%  \\
    \bottomrule
    \end{tabular}
    \vspace{5pt}
    \caption{Human preferences on images generated by DCGAN~\citep{Radford2016Unsupervised}, LSGAN~\citep{mao2017least}, WGAN-GP~\citep{Gulrajani2017Improved} \vs our \oursshort (DF). Each number shows the percentage of human subjects who prefer our method to the baseline method. }
    \label{tbl:user2d}
    \vspace{-20pt}
    }
\end{table}

%% file: figText/tbl_fid3d.tex
\begin{minipage}{\textwidth}
\begin{minipage}{0.45\textwidth}
\includegraphics[width=\linewidth]{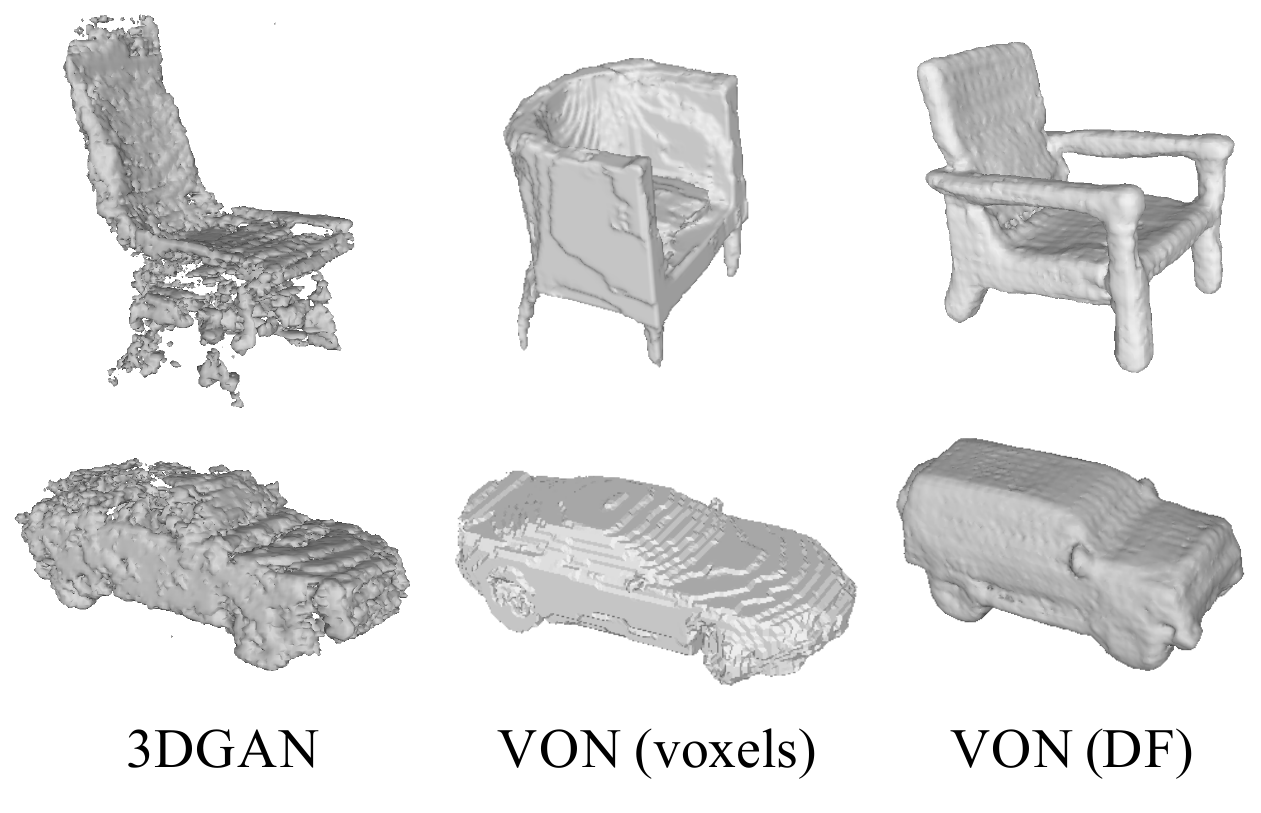}
\vspace{-18pt}
\captionof{figure}{Sampled 3D shapes: from left to right: 3DGAN~\citep{Wu2016Learning},  \oursshort on voxels, VON on distance functions (DF). Our model produce more natural 3D shapes. OUr model produces samples with higher quality.}
\lblfig{shape_comp}
\end{minipage}
\hspace{3pt}
\begin{minipage}{0.5\textwidth}
    \begin{tabular}{lcc}
    \\
    \toprule
     & 3D-GAN (voxels)  &  \oursshort (voxels) \\
    \midrule
    Cars &3.021 &  {\bf 0.021} \\
    Chairs & 2.598 & {\bf 0.082} \\
    \toprule
     & 3D-GAN (DF) & \oursshort (DF)  \\
    \midrule
    Cars & 3.896 & {\bf 0.002}  \\
    Chairs & 1.790 & {\bf 0.006}   \\
    \bottomrule
    \end{tabular}
    \vspace{10pt}
    \captionof{table}{Quantitative comparisons on shape generation: \fids (FID) between real shapes and shapes generated by 3D-GAN~\citep{Wu2016Learning} and our shape network, both on voxels and distance function representation (DF). Our model achieves better results regarding FID. }
    \lbltbl{fid3d}
\end{minipage}
\end{minipage}

%% file: figText/applications.tex
\begin{figure}[t]
  \centering
     \includegraphics[width=\linewidth]{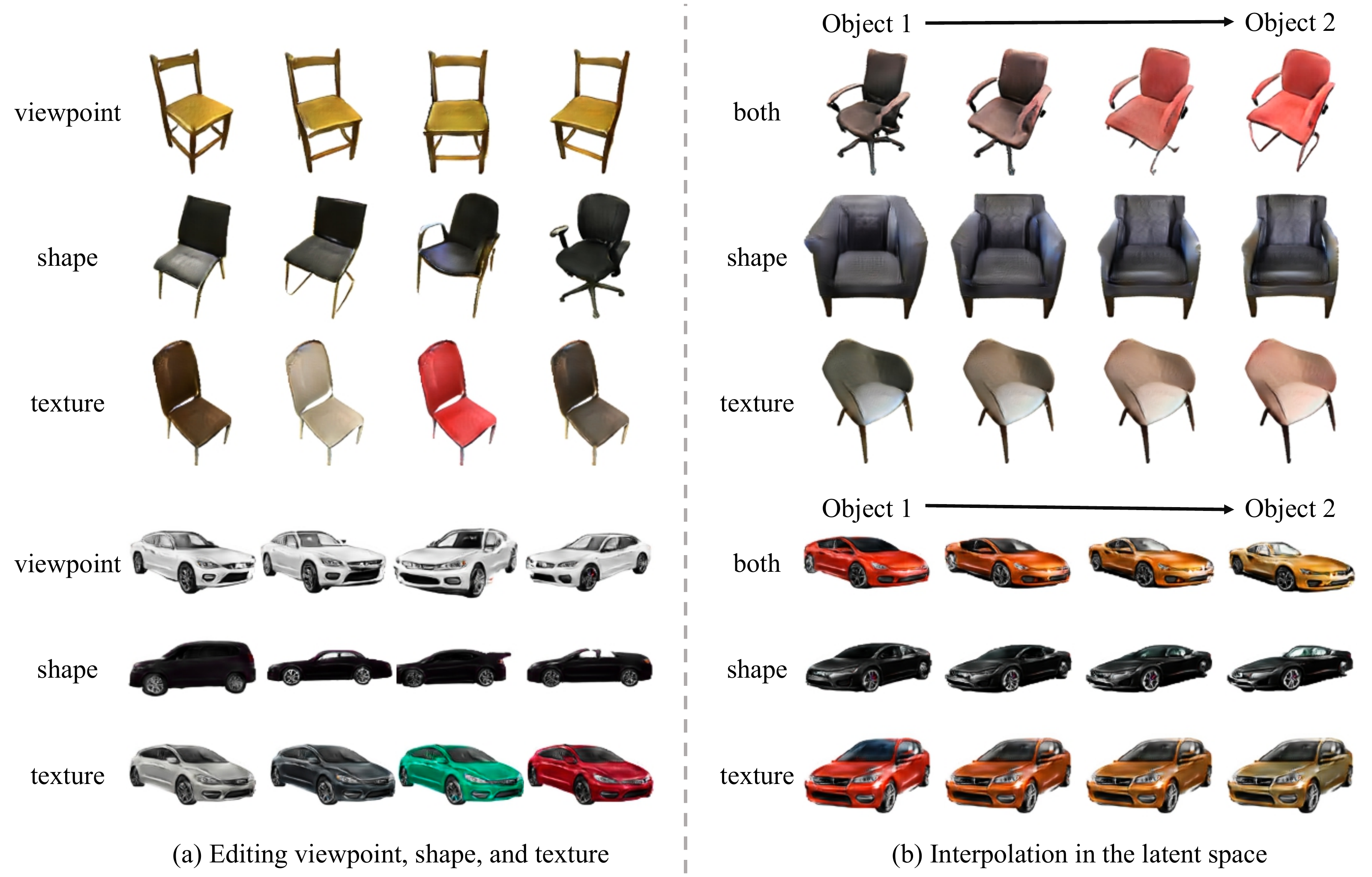}
  \caption{3D-aware applications:  Our \ours allow several 3D applications such as (a) changing the viewpoint, texture, or shape independently, and (b) interpolating between two objects in shape space, texture space, or both. None of them can be achieved by previous 2D GANs.}
  \lblfig{app}
  \vspace{-5pt}
\end{figure}

%% file: figText/transfer.tex
\begin{figure}[t]
  \centering
   \includegraphics[width=0.9\linewidth]{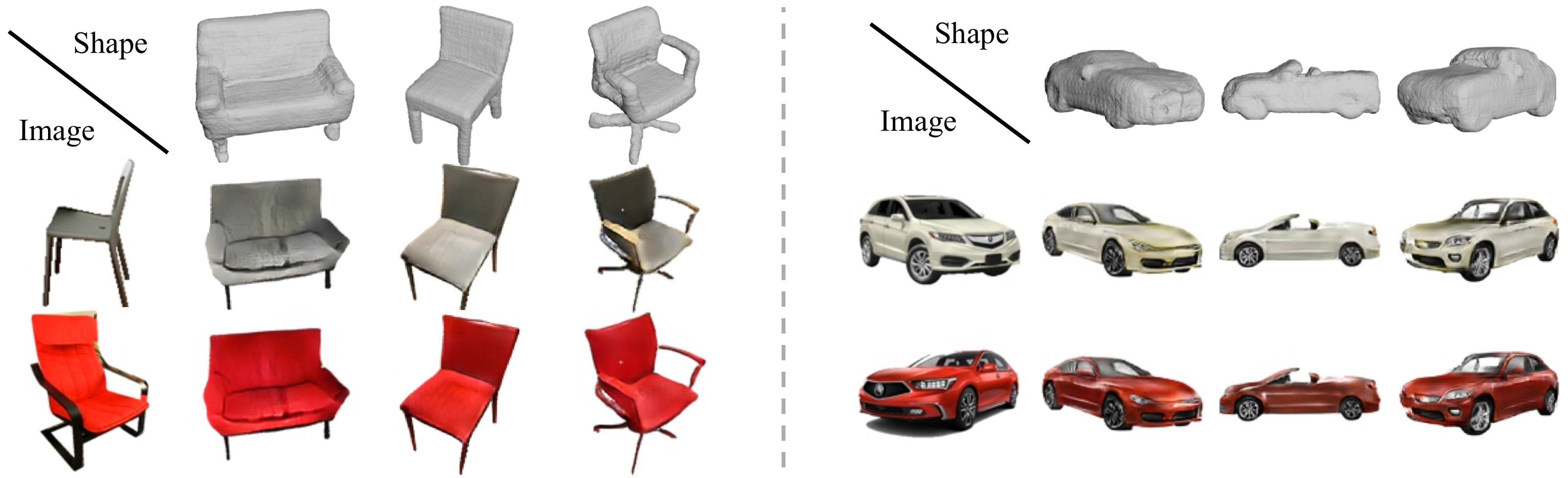}
     \vspace{-5pt}
  \caption{Given a real input image, we synthesize objects with similar texture using the inferred texture code. The same texture is transferred to different shapes and viewpoints.}
  \lblfig{transfer}
  \vspace{-10pt}
\end{figure}

%% file: text/conclusion.tex
\section{Discussion}
\vspace{-8pt}
\lblsec{conclusion}
In this paper, we have presented \ours (\oursshorts), a fully differentiable 3D-aware generative model for image and shape synthesis. Our key idea is to disentangle the image generation process into three factors: shape, viewpoint, and texture. This disentangled 3D representation allows us to learn the model from both 3D and 2D visual data collections under an adversarial learning framework. Our model synthesizes more photorealistic images compared to existing 2D generative models; it also enables various 3D manipulations that are not possible with prior 2D methods. 

In the future, we are interested in incorporating coarse-to-fine modeling~\citep{Karras2017Progressive} for producing shapes and images at a higher resolution. Another interesting direction to explore is to disentangle texture further into lighting and appearance (e.g., albedo), which could improve the consistency of appearance across different viewpoints and lighting conditions. Finally, as we do not have large-scale 3D geometric data for entire scenes, our current method only works for individual objects. Synthesizing natural scenes is also a meaningful next step.

\myparagraph{Acknowledgements}
This work is supported by NSF \#1231216, NSF \#1524817, ONR MURI N00014-16-1-2007, Toyota Research Institute, Shell, and Facebook. We thank Xiuming Zhang, Richard Zhang, David Bau, and Zhuang Liu for valuable discussions.

%% file: main.bbl
\begin{thebibliography}{63}
\providecommand{\natexlab}[1]{#1}
\providecommand{\url}[1]{\texttt{#1}}
\expandafter\ifx\csname urlstyle\endcsname\relax
  \providecommand{\doi}[1]{doi: #1}\else
  \providecommand{\doi}{doi: \begingroup \urlstyle{rm}\Url}\fi

\bibitem[Achlioptas et~al.(2018)Achlioptas, Diamanti, Mitliagkas, and
  Guibas]{Achlioptas2018Learning}
Panos Achlioptas, Olga Diamanti, Ioannis Mitliagkas, and Leonidas Guibas.
\newblock Learning representations and generative models for 3d point clouds.
\newblock In \emph{ICLR Workshop}, 2018.

\bibitem[Almahairi et~al.(2018)Almahairi, Rajeswar, Sordoni, Bachman, and
  Courville]{almahairi2018augmented}
Amjad Almahairi, Sai Rajeswar, Alessandro Sordoni, Philip Bachman, and Aaron
  Courville.
\newblock Augmented cyclegan: Learning many-to-many mappings from unpaired
  data.
\newblock In \emph{ICML}, 2018.

\bibitem[Arjovsky et~al.(2017)Arjovsky, Chintala, and
  Bottou]{arjovsky2017wasserstein}
Mart{\'{\i}}n Arjovsky, Soumith Chintala, and L{\'{e}}on Bottou.
\newblock Wasserstein generative adversarial networks.
\newblock In \emph{ICML}, 2017.

\bibitem[Barrow and Tenenbaum(1978)]{Barrow1978Recovering}
Harry~G Barrow and Jay~M Tenenbaum.
\newblock Recovering intrinsic scene characteristics from images.
\newblock \emph{Computer Vision Systems}, 1978.

\bibitem[Bever and Poeppel(2010)]{Bever2010Analysis}
Thomas~G Bever and David Poeppel.
\newblock Analysis by synthesis: a (re-) emerging program of research for
  language and vision.
\newblock \emph{Biolinguistics}, 4\penalty0 (2-3):\penalty0 174--200, 2010.

\bibitem[Blanz and Vetter(1999)]{blanz1999morphable}
Volker Blanz and Thomas Vetter.
\newblock A morphable model for the synthesis of 3d faces.
\newblock In \emph{SIGGRAPH}, 1999.

\bibitem[Chang et~al.(2015)Chang, Funkhouser, Guibas, Hanrahan, Huang, Li,
  Savarese, Savva, Song, Su, Xiao, Yi, and Yu]{Chang2015Shapenet}
Angel~X Chang, Thomas Funkhouser, Leonidas Guibas, Pat Hanrahan, Qixing Huang,
  Zimo Li, Silvio Savarese, Manolis Savva, Shuran Song, Hao Su, Jianxiong Xiao,
  Li~Yi, and Fisher Yu.
\newblock {Shapenet: An information-rich 3d model repository}.
\newblock \emph{arXiv:1512.03012}, 2015.

\bibitem[Chen et~al.(2016)Chen, Duan, Houthooft, Schulman, Sutskever, and
  Abbeel]{chen2016InfoGAN}
Xi~Chen, Yan Duan, Rein Houthooft, John Schulman, Ilya Sutskever, and Pieter
  Abbeel.
\newblock Infogan: Interpretable representation learning by information
  maximizing generative adversarial nets.
\newblock In \emph{NIPS}, 2016.

\bibitem[Curless and Levoy(1996)]{curless1996volumetric}
Brian Curless and Marc Levoy.
\newblock A volumetric method for building complex models from range images.
\newblock In \emph{SIGGRAPH}, 1996.

\bibitem[Dai et~al.(2017)Dai, Qi, and Nie{\ss}ner]{Dai2017Shape}
Angela Dai, Charles~Ruizhongtai Qi, and Matthias Nie{\ss}ner.
\newblock Shape completion using 3d-encoder-predictor cnns and shape synthesis.
\newblock In \emph{CVPR}, 2017.

\bibitem[Deng et~al.(2009)Deng, Dong, Socher, Li, Li, and
  Fei-Fei]{Deng2009Imagenet}
Jia Deng, Wei Dong, Richard Socher, Li-Jia Li, Kai Li, and Li~Fei-Fei.
\newblock Imagenet: A large-scale hierarchical image database.
\newblock In \emph{CVPR}, 2009.

\bibitem[Dosovitskiy et~al.(2015)Dosovitskiy, Tobias~Springenberg, and
  Brox]{Dosovitskiy2015Learning}
Alexey Dosovitskiy, Jost Tobias~Springenberg, and Thomas Brox.
\newblock {Learning to generate chairs with convolutional neural networks}.
\newblock In \emph{CVPR}, 2015.

\bibitem[Gadelha et~al.(2017{\natexlab{a}})Gadelha, Maji, and
  Wang]{gadelha20173d}
Matheus Gadelha, Subhransu Maji, and Rui Wang.
\newblock 3d shape induction from 2d views of multiple objects.
\newblock In \emph{3D Vision (3DV)}, pages 402--411. IEEE, 2017{\natexlab{a}}.

\bibitem[Gadelha et~al.(2017{\natexlab{b}})Gadelha, Maji, and
  Wang]{gadelha2017shape}
Matheus Gadelha, Subhransu Maji, and Rui Wang.
\newblock Shape generation using spatially partitioned point clouds.
\newblock In \emph{BMVC}, 2017{\natexlab{b}}.

\bibitem[Goodfellow et~al.(2014)Goodfellow, Pouget-Abadie, Mirza, Xu,
  Warde-Farley, Ozair, Courville, and Bengio]{Goodfellow2014Generative}
Ian Goodfellow, Jean Pouget-Abadie, Mehdi Mirza, Bing Xu, David Warde-Farley,
  Sherjil Ozair, Aaron Courville, and Yoshua Bengio.
\newblock Generative adversarial nets.
\newblock In \emph{NIPS}, 2014.

\bibitem[Gulrajani et~al.(2017)Gulrajani, Ahmed, Arjovsky, Dumoulin, and
  Courville]{Gulrajani2017Improved}
Ishaan Gulrajani, Faruk Ahmed, Martin Arjovsky, Vincent Dumoulin, and Aaron
  Courville.
\newblock Improved training of wasserstein gans.
\newblock In \emph{NIPS}, 2017.

\bibitem[He et~al.(2015)He, Zhang, Ren, and Sun]{He2015Deep}
Kaiming He, Xiangyu Zhang, Shaoqing Ren, and Jian Sun.
\newblock Deep residual learning for image recognition.
\newblock In \emph{CVPR}, 2015.

\bibitem[Heusel et~al.(2017)Heusel, Ramsauer, Unterthiner, Nessler, Klambauer,
  and Hochreiter]{heusel2017gans}
Martin Heusel, Hubert Ramsauer, Thomas Unterthiner, Bernhard Nessler,
  G{\"u}nter Klambauer, and Sepp Hochreiter.
\newblock Gans trained by a two time-scale update rule converge to a nash
  equilibrium.
\newblock \emph{NIPS1706.08500}, 2017.

\bibitem[Huang et~al.(2018)Huang, Liu, Belongie, and
  Kautz]{Huang2018Multimodal}
Xun Huang, Ming-Yu Liu, Serge Belongie, and Jan Kautz.
\newblock Multimodal unsupervised image-to-image translation.
\newblock \emph{ECCV1804.04732}, 2018.

\bibitem[Ioffe and Szegedy(2015)]{ioffe2015batch}
Sergey Ioffe and Christian Szegedy.
\newblock Batch normalization: Accelerating deep network training by reducing
  internal covariate shift.
\newblock In \emph{ICML}, 2015.

\bibitem[Isola et~al.(2016)Isola, Zoran, Krishnan, and
  Adelson]{Isola2016Learning}
Phillip Isola, Daniel Zoran, Dilip Krishnan, and Edward~H Adelson.
\newblock Learning visual groups from co-occurrences in space and time.
\newblock In \emph{ICLR Workshop}, 2016.

\bibitem[Isola et~al.(2017)Isola, Zhu, Zhou, and Efros]{Isola2017Image}
Phillip Isola, Jun-Yan Zhu, Tinghui Zhou, and Alexei~A Efros.
\newblock Image-to-image translation with conditional adversarial networks.
\newblock In \emph{CVPR}, 2017.

\bibitem[Jaderberg et~al.(2015)Jaderberg, Simonyan, and
  Zisserman]{jaderberg2015spatial}
Max Jaderberg, Karen Simonyan, and Andrew Zisserman.
\newblock Spatial transformer networks.
\newblock In \emph{NIPS}, 2015.

\bibitem[Kajiya(1986)]{kajiya1986rendering}
James~T Kajiya.
\newblock The rendering equation.
\newblock In \emph{SIGGRAPH}, 1986.

\bibitem[Kanazawa et~al.(2018)Kanazawa, Tulsiani, Efros, and
  Malik]{Kanazawa2018Learning}
Angjoo Kanazawa, Shubham Tulsiani, Alexei~A Efros, and Jitendra Malik.
\newblock Learning category-specific mesh reconstruction from image
  collections.
\newblock \emph{ECCV1803.07549}, 2018.

\bibitem[Karras et~al.(2017)Karras, Aila, Laine, and
  Lehtinen]{Karras2017Progressive}
Tero Karras, Timo Aila, Samuli Laine, and Jaakko Lehtinen.
\newblock Progressive growing of gans for improved quality, stability, and
  variation.
\newblock In \emph{ICLR}, 2017.

\bibitem[Karras et~al.(2018)Karras, Aila, Laine, and
  Lehtinen]{karras2018progressive}
Tero Karras, Timo Aila, Samuli Laine, and Jaakko Lehtinen.
\newblock Progressive growing of gans for improved quality, stability, and
  variation.
\newblock \emph{ICLR}, 2018.

\bibitem[Kingma and Ba(2015)]{Kingma2015Adam}
Diederik~P. Kingma and Jimmy Ba.
\newblock Adam: A method for stochastic optimization.
\newblock In \emph{ICLR}, 2015.

\bibitem[Kingma and Welling(2014)]{Kingma2014Auto}
Diederik~P. Kingma and Max Welling.
\newblock Auto-encoding variational bayes.
\newblock In \emph{ICLR}, 2014.

\bibitem[Kulkarni et~al.(2015{\natexlab{a}})Kulkarni, Kohli, Tenenbaum, and
  Mansinghka]{Kulkarni2015Picture}
Tejas~D Kulkarni, Pushmeet Kohli, Joshua~B Tenenbaum, and Vikash Mansinghka.
\newblock Picture: A probabilistic programming language for scene perception.
\newblock In \emph{CVPR}, 2015{\natexlab{a}}.

\bibitem[Kulkarni et~al.(2015{\natexlab{b}})Kulkarni, Whitney, Kohli, and
  Tenenbaum]{Kulkarni2015Deep}
Tejas~D Kulkarni, William~F Whitney, Pushmeet Kohli, and Josh Tenenbaum.
\newblock Deep convolutional inverse graphics network.
\newblock In \emph{NIPS}, 2015{\natexlab{b}}.

\bibitem[Ledig et~al.(2017)Ledig, Theis, Husz{\'a}r, Caballero, Cunningham,
  Acosta, Aitken, Tejani, Totz, Wang, and Shi]{Ledig2017Photo}
Christian Ledig, Lucas Theis, Ferenc Husz{\'a}r, Jose Caballero, Andrew
  Cunningham, Alejandro Acosta, Andrew Aitken, Alykhan Tejani, Johannes Totz,
  Zehan Wang, and Wenzhe Shi.
\newblock Photo-realistic single image super-resolution using a generative
  adversarial network.
\newblock In \emph{CVPR}, 2017.

\bibitem[Liu et~al.(2017)Liu, Breuel, and Kautz]{liu2017unsupervised}
Ming-Yu Liu, Thomas Breuel, and Jan Kautz.
\newblock Unsupervised image-to-image translation networks.
\newblock In \emph{NIPS}, 2017.

\bibitem[Lucic et~al.(2018)Lucic, Kurach, Michalski, Gelly, and
  Bousquet]{lucic2017gans}
Mario Lucic, Karol Kurach, Marcin Michalski, Sylvain Gelly, and Olivier
  Bousquet.
\newblock Are gans created equal? a large-scale study.
\newblock \emph{NIPS}, 2018.

\bibitem[Mao et~al.(2017)Mao, Li, Xie, Lau, Wang, and Smolley]{mao2017least}
Xudong Mao, Qing Li, Haoran Xie, Raymond~YK Lau, Zhen Wang, and Stephen~Paul
  Smolley.
\newblock Least squares generative adversarial networks.
\newblock In \emph{ICCV}, 2017.

\bibitem[Marr(1982)]{Marr1982Vision}
David Marr.
\newblock \emph{Vision: A computational investigation into the human
  representation and processing of visual information}, volume~2.
\newblock W. H. Freeman and Company, 1982.

\bibitem[Mathieu et~al.(2016)Mathieu, Couprie, and LeCun]{Mathieu2016Deep}
Michael Mathieu, Camille Couprie, and Yann LeCun.
\newblock Deep multi-scale video prediction beyond mean square error.
\newblock In \emph{ICLR}, 2016.

\bibitem[Pathak et~al.(2016)Pathak, Krahenbuhl, Donahue, Darrell, and
  Efros]{pathak2016context}
Deepak Pathak, Philipp Krahenbuhl, Jeff Donahue, Trevor Darrell, and Alexei~A
  Efros.
\newblock Context encoders: Feature learning by inpainting.
\newblock In \emph{CVPR}, 2016.

\bibitem[Radford et~al.(2016)Radford, Metz, and
  Chintala]{Radford2016Unsupervised}
Alec Radford, Luke Metz, and Soumith Chintala.
\newblock Unsupervised representation learning with deep convolutional
  generative adversarial networks.
\newblock In \emph{ICLR}, 2016.

\bibitem[Reed et~al.(2016)Reed, Akata, Yan, Logeswaran, Schiele, and
  Lee]{reed2016generative}
Scott Reed, Zeynep Akata, Xinchen Yan, Lajanugen Logeswaran, Bernt Schiele, and
  Honglak Lee.
\newblock Generative adversarial text-to-image synthesis.
\newblock In \emph{ICML}, 2016.

\bibitem[Rezende et~al.(2016)Rezende, Eslami, Mohamed, Battaglia, Jaderberg,
  and Heess]{Rezende2016Unsupervised}
Danilo~Jimenez Rezende, SM~Eslami, Shakir Mohamed, Peter Battaglia, Max
  Jaderberg, and Nicolas Heess.
\newblock Unsupervised learning of 3d structure from images.
\newblock In \emph{NIPS}, 2016.

\bibitem[Shu et~al.(2017)Shu, Yumer, Hadap, Sunkavalli, Shechtman, and
  Samaras]{shu2017neural}
Zhixin Shu, Ersin Yumer, Sunil Hadap, Kalyan Sunkavalli, Eli Shechtman, and
  Dimitris Samaras.
\newblock Neural face editing with intrinsic image disentangling.
\newblock In \emph{CVPR}, 2017.

\bibitem[Sun et~al.(2018{\natexlab{a}})Sun, Wu, Zhang, Zhang, Zhang, Xue,
  Tenenbaum, and Freeman]{sun2018pix3d}
Xingyuan Sun, Jiajun Wu, Xiuming Zhang, Zhoutong Zhang, Chengkai Zhang, Tianfan
  Xue, Joshua~B Tenenbaum, and William~T Freeman.
\newblock Pix3d: Dataset and methods for single-image 3d shape modeling.
\newblock In \emph{CVPR}, 2018{\natexlab{a}}.

\bibitem[Sun et~al.(2018{\natexlab{b}})Sun, Liu, Wang, and
  Sarma]{Sun2018Im2Avatar}
Yongbin Sun, Ziwei Liu, Yue Wang, and Sanjay~E Sarma.
\newblock Im2avatar: Colorful 3d reconstruction from a single image.
\newblock \emph{arXiv:1804.06375}, 2018{\natexlab{b}}.

\bibitem[Szegedy et~al.(2015)Szegedy, Liu, Jia, Sermanet, Reed, Anguelov,
  Erhan, Vanhoucke, and Rabinovich]{Szegedy2015Going}
Christian Szegedy, Wei Liu, Yangqing Jia, Pierre Sermanet, Scott Reed, Dragomir
  Anguelov, Dumitru Erhan, Vincent Vanhoucke, and Andrew Rabinovich.
\newblock Going deeper with convolutions.
\newblock In \emph{CVPR}, 2015.

\bibitem[Tatarchenko et~al.(2016)Tatarchenko, Dosovitskiy, and
  Brox]{Tatarchenko2016Multi}
Maxim Tatarchenko, Alexey Dosovitskiy, and Thomas Brox.
\newblock Multi-view 3d models from single images with a convolutional network.
\newblock In \emph{ECCV}, 2016.

\bibitem[Tatarchenko et~al.(2017)Tatarchenko, Dosovitskiy, and
  Brox]{Tatarchenko2017Octree}
Maxim Tatarchenko, Alexey Dosovitskiy, and Thomas Brox.
\newblock Octree generating networks: Efficient convolutional architectures for
  high-resolution 3d outputs.
\newblock In \emph{ICCV}, 2017.

\bibitem[Tulsiani et~al.(2017)Tulsiani, Su, Guibas, Efros, and
  Malik]{Tulsiani2017Learning}
Shubham Tulsiani, Hao Su, Leonidas~J Guibas, Alexei~A Efros, and Jitendra
  Malik.
\newblock Learning shape abstractions by assembling volumetric primitives.
\newblock In \emph{CVPR}, 2017.

\bibitem[Tung et~al.(2017)Tung, Harley, Seto, and
  Fragkiadaki]{tung2017adversarial}
Hsiao-Yu~Fish Tung, Adam~W Harley, William Seto, and Katerina Fragkiadaki.
\newblock Adversarial inverse graphics networks: Learning 2d-to-3d lifting and
  image-to-image translation from unpaired supervision.
\newblock In \emph{ICCV}, 2017.

\bibitem[Ulyanov et~al.(2016)Ulyanov, Vedaldi, and
  Lempitsky]{Ulyanov2016Instance}
Dmitry Ulyanov, Andrea Vedaldi, and Victor~S. Lempitsky.
\newblock Instance normalization: The missing ingredient for fast stylization.
\newblock \emph{arXiv:1607.08022}, 2016.

\bibitem[Wang et~al.(2018)Wang, Liu, Zhu, Tao, Kautz, and
  Catanzaro]{wang2018high}
Ting-Chun Wang, Ming-Yu Liu, Jun-Yan Zhu, Andrew Tao, Jan Kautz, and Bryan
  Catanzaro.
\newblock High-resolution image synthesis and semantic manipulation with
  conditional gans.
\newblock In \emph{CVPR}, 2018.

\bibitem[Wang et~al.(2017)Wang, Huang, You, Yang, and Neumann]{Wang2018Shape}
Weiyue Wang, Qiangui Huang, Suya You, Chao Yang, and Ulrich Neumann.
\newblock Shape inpainting using 3d generative adversarial network and
  recurrent convolutional networks.
\newblock In \emph{ICCV}, 2017.

\bibitem[Wu et~al.(2016)Wu, Zhang, Xue, Freeman, and Tenenbaum]{Wu2016Learning}
Jiajun Wu, Chengkai Zhang, Tianfan Xue, William~T Freeman, and Joshua~B
  Tenenbaum.
\newblock {Learning a Probabilistic Latent Space of Object Shapes via 3D
  Generative-Adversarial Modeling}.
\newblock In \emph{NIPS}, 2016.

\bibitem[Wu et~al.(2017)Wu, Wang, Xue, Sun, Freeman, and
  Tenenbaum]{Wu2017MarrNet}
Jiajun Wu, Yifan Wang, Tianfan Xue, Xingyuan Sun, William~T Freeman, and
  Joshua~B Tenenbaum.
\newblock {MarrNet: 3D Shape Reconstruction via 2.5D Sketches}.
\newblock In \emph{NIPS}, 2017.

\bibitem[Wu et~al.(2018)Wu, Zhang, Zhang, Zhang, Freeman, and
  Tenenbaum]{wu2018learning}
Jiajun Wu, Chengkai Zhang, Xiuming Zhang, Zhoutong Zhang, William~T Freeman,
  and Joshua~B Tenenbaum.
\newblock Learning 3d shape priors for shape completion and reconstruction.
\newblock In \emph{ECCV}, 2018.

\bibitem[Yang et~al.(2015)Yang, Reed, Yang, and Lee]{Yang2015Weakly}
Jimei Yang, Scott~E Reed, Ming-Hsuan Yang, and Honglak Lee.
\newblock Weakly-supervised disentangling with recurrent transformations for 3d
  view synthesis.
\newblock In \emph{NIPS}, 2015.

\bibitem[Yi et~al.(2017)Yi, Zhang, Tan, and Gong]{yi2017dualgan}
Zili Yi, Hao~(Richard) Zhang, Ping Tan, and Minglun Gong.
\newblock Dualgan: Unsupervised dual learning for image-to-image translation.
\newblock In \emph{ICCV}, 2017.

\bibitem[Yuille and Kersten(2006)]{Yuille2006Vision}
Alan Yuille and Daniel Kersten.
\newblock Vision as bayesian inference: analysis by synthesis?
\newblock \emph{TiCS}, 10\penalty0 (7):\penalty0 301--308, 2006.

\bibitem[Zhang et~al.(2017)Zhang, Xu, Li, Zhang, Wang, Huang, and
  Metaxas]{zhang2017stackgan}
Han Zhang, Tao Xu, Hongsheng Li, Shaoting Zhang, Xiaogang Wang, Xiaolei Huang,
  and Dimitris Metaxas.
\newblock Stackgan: Text to photo-realistic image synthesis with stacked
  generative adversarial networks.
\newblock In \emph{ICCV}, 2017.

\bibitem[Zhang et~al.(2018)Zhang, Zhang, Zhang, Tenenbaum, Freeman, and
  Wu]{Zhang2018Learning}
Xiuming Zhang, Zhoutong Zhang, Chengkai Zhang, Joshua~B Tenenbaum, William~T
  Freeman, and Jiajun Wu.
\newblock Learning to reconstruct shapes from unseen categories.
\newblock In \emph{NIPS}, 2018.

\bibitem[Zhu et~al.(2016)Zhu, Kr{\"a}henb{\"u}hl, Shechtman, and
  Efros]{Zhu2016Generative}
Jun-Yan Zhu, Philipp Kr{\"a}henb{\"u}hl, Eli Shechtman, and Alexei~A Efros.
\newblock Generative visual manipulation on the natural image manifold.
\newblock In \emph{ECCV}, 2016.

\bibitem[Zhu et~al.(2017{\natexlab{a}})Zhu, Park, Isola, and
  Efros]{Zhu2017Unpaired}
Jun-Yan Zhu, Taesung Park, Phillip Isola, and Alexei~A Efros.
\newblock Unpaired image-to-image translation using cycle-consistent
  adversarial networks.
\newblock In \emph{ICCV}, 2017{\natexlab{a}}.

\bibitem[Zhu et~al.(2017{\natexlab{b}})Zhu, Zhang, Pathak, Darrell, Efros,
  Wang, and Shechtman]{Zhu2017Toward}
Jun-Yan Zhu, Richard Zhang, Deepak Pathak, Trevor Darrell, Alexei~A Efros,
  Oliver Wang, and Eli Shechtman.
\newblock Toward multimodal image-to-image translation.
\newblock In \emph{NIPS}, 2017{\natexlab{b}}.

\end{thebibliography}
